\documentclass{article}

\usepackage[utf8]{inputenc}
\usepackage{amssymb}
\usepackage{microtype}
\usepackage{graphicx}
\usepackage{subcaption}
\usepackage{booktabs} 
\usepackage{multirow}
\usepackage{multicol}
\usepackage{array}

\usepackage[table]{xcolor} 
\usepackage{hyperref}

\usepackage[preprint]{icml2026}

\usepackage{amsmath}
\usepackage{amssymb}
\usepackage{mathtools}
\usepackage{amsthm}
\usepackage{tcolorbox}
\usepackage{fancyvrb}
\usepackage{listings}
\tcbuselibrary{listings,skins,breakable}

\definecolor{codegreen}{rgb}{0,0.6,0}
\definecolor{codegray}{rgb}{0.5,0.5,0.5}
\definecolor{codepurple}{rgb}{0.58,0,0.82}
\definecolor{backcolour}{rgb}{0.91,0.91,0.9}
\definecolor{shallowRed}{rgb}{1, 0.8, 0.8}
\definecolor{shallowYellow}{rgb}{1, 0.953, 0.8}
\definecolor{shallowBlue}{rgb}{0.8, 0.8, 1}
\definecolor{orange}{rgb}{1, 0.6, 0}
\definecolor{shallowOrange}{rgb}{1, 0.88, 0.7}

\definecolor{keywordcolor}{rgb}{0.7, 0.1, 0.1}   
\definecolor{commentcolor}{rgb}{0.4, 0.4, 0.4}   
\definecolor{symbolcolor}{rgb}{0.0, 0.1, 0.6}    
\definecolor{sortcolor}{rgb}{0.1, 0.5, 0.1}      
\definecolor{errorcolor}{rgb}{1, 0, 0}           
\definecolor{stringcolor}{rgb}{0.5, 0.3, 0.2}    

\definecolor{leftcolor}{rgb}{0.522, 0.765, 0.863}  
\definecolor{midcolor}{rgb}{0.855, 0.776, 0.812}
\definecolor{rightcolor}{rgb}{0.886, 0.753, 0.596}

\usepackage[capitalize,noabbrev]{cleveref}

\theoremstyle{plain}

\theoremstyle{definition}

\theoremstyle{remark}

\usepackage[textsize=tiny]{todonotes}
\usepackage{enumitem}

\begin{document}

\twocolumn[
  \icmltitle{OPE: Overcoming Information Saturation in Parallel Thinking via Outline-Guided Path Exploration}



  \icmlsetsymbol{corres}{†}

  \begin{icmlauthorlist}
    \icmlauthor{Qi Guo}{pku,meituan}
    \icmlauthor{Jianing Wang}{corres,meituan}
    \icmlauthor{Deyang Kong}{pku,meituan}
    \icmlauthor{Xiangyu Xi}{meituan}
    \icmlauthor{Jianfei Zhang}{meituan}
    \icmlauthor{Yi Lu}{meituan}
    \icmlauthor{Jingang Wang}{meituan}
    \icmlauthor{Wei Wang}{meituan}
    \icmlauthor{Shikun Zhang}{pku}
    \icmlauthor{Wei Ye}{corres,pku}
  \end{icmlauthorlist}

  \icmlaffiliation{pku}{National Engineering Research Center for Software Engineering, Peking University, Beijing, China}
  \icmlaffiliation{meituan}{Meituan Group, Beijing, China}

  \icmlcorrespondingauthor{Jianing Wang}{lygwjn@gmail.com}
  \icmlcorrespondingauthor{Wei Ye}{wye@pku.edu.cn}

  \icmlkeywords{Machine Learning, ICML}

  \vskip 0.3in
]



\printAffiliationsAndNotice{}  

\begin{abstract}
Parallel thinking has emerged as a new paradigm for large reasoning models (LRMs) in tackling complex problems. Recent methods leverage Reinforcement Learning (RL) to enhance parallel thinking, aiming to address the limitations in computational resources and effectiveness encountered with supervised fine-tuning. However, most existing studies primarily focus on optimizing the aggregation phase, with limited attention to the path exploration stage.
In this paper, we theoretically analyze the optimization of parallel thinking under the Reinforcement Learning with Verifiable Rewards (RLVR) setting, and identify that the mutual information bottleneck among exploration paths fundamentally restricts overall performance. 
To address this, we propose Outline-Guided Path Exploration (OPE), which explicitly partitions the solution space by generating diverse reasoning outlines prior to parallel path reasoning, thereby reducing information redundancy and improving the diversity of information captured across exploration paths. 
We implement OPE with an iterative RL strategy that optimizes outline planning and outline-guided reasoning independently. Extensive experiments across multiple challenging mathematical benchmarks demonstrate that OPE effectively improves reasoning performance in different aggregation strategies, enabling LRMs to more reliably discover correct solutions.
\vskip -0.8in
\end{abstract}

\section{Introduction}
Despite the rapid evolution of Large Reasoning Models (LRMs) \citep{deepseek-r1, longcat, qwen3}, their capacity to solve complex problems in a single inference pass remains constrained. 
Following the success of Gemini \citep{gemini3} in the International Mathematical Olympiad competitions, parallel thinking has emerged as a promising reasoning paradigm for addressing such challenges. 
Parallel thinking allows the model to concurrently explore multiple potential solution paths, subsequently aggregating the outcomes of these explorations to synthesize the final answer. 
Such a paradigm encourages the model to broadly search for potential correct trajectories within the solution space, while simultaneously mitigating the risk of biased outcomes associated with a single exploration trajectory \citep{adaptive_parallel, parallel-r1}.
\par
Prior approaches rely on Supervised Fine-Tuning (SFT) over extensive corpus of high-quality parallel reasoning trajectories \citep{parathinker, multiverse, sot}. 
This process not only incurs heavy resource consumption for trajectory synthesis but also risks superficial pattern imitation, thereby limiting effectiveness on challenging tasks.
Consequently, recent studies have shifted toward utilizing Reinforcement Learning (RL) to empower models to autonomously acquire parallel thinking under practical problem-solving scenarios \citep{parallel-r1}. 
However, the application of RL in parallel thinking remains underexplored, with existing works predominantly concentrating on the aggregation phase over the exploration phase. 
Crucially, prior studies indicate that the effectiveness of the aggregation phase depends heavily on the quality of the paths found during exploration, which sets a natural limit on overall performance of parallel thinking \citep{wang2025survey}.
\par
In this paper, we investigate the impact of the path exploration stage on the performance of parallel thinking. 
We present a theoretical analysis within the Reinforcement Learning with Verifiable Rewards (RLVR), and reveal that overall performance is fundamentally constrained by mutual information saturation among explored paths. 
Specifically, under the naive parallel thinking paradigm where paths are sampled independently, optimizing the information gain between trajectories becomes challenging. 
In addition, since LRMs tend to suffer from mode collapse, the generated paths often exhibit high redundancy \cite{curse, shao2024deepseekmath}. 
Consequently, when tackling complex problems, multiple parallel paths often converge on the same incorrect answers rather than covering diverse regions of the solution space.
Based on these findings, we propose Outline-guided Parallel Exploration (OPE), a framework that requires the model to explicitly partition the solution space using outlines before path reasoning, thereby maximizing the mutual information between generated paths and the correct solution. 
We start from a cold-start stage to instill outline planning capabilities, where the model learns to analyze problems and generate outlines for diverse reasoning strategies. 
Building on this, we introduce an iterative RL strategy to jointly optimize outline planning and outline-guided reasoning capabilities. 
Specifically, we first conduct Outline Planning RL to encourage the model to generate diverse outlines that define multiple distinct problem-solving directions. This is followed by Path Reasoning RL, which refines the model's ability to execute reasoning steps under the guidance of these outlines. These two training phases are interleaved at fixed step intervals to facilitate the synergistic evolution of both capabilities.
\par
Experimental results across serveral mathematical reasoning benchmarks demonstrate that OPE leads to significant improvements in parallel thinking performance, with particularly substantial gains on the most challenging tasks.
Furthermore, comparative analysis shows that, compared to naive parallel thinking which samples each path independently, OPE exhibits superior scaling properties at test time and alleviates the issue of ``overthinking" during path reasoning, enabling the model to reach solutions more efficiently through targeted exploration.
\par
The contributions of this paper can be summarized as:
\begin{itemize}[itemsep=8pt, parsep=2pt, topsep=0pt, partopsep=0pt]
\item We provide a theoretical perspective on parallel thinking RL, and formally identify mutual information saturation during the exploration phase as the fundamental bottleneck limiting performance.
\item We introduce Outline-guided Parallel Exploration (OPE), a novel paradigm designed to mitigate mutual information saturation by explicitly partitioning the solution space with outlines. We implement OPE via a novel iterative RL strategy that synergistically optimizes outline planning and path reasoning capabilities.
\item Extensive experiments demonstrate that OPE achieves superior performance across complex benchmarks. Further analysis reveals that OPE exhibits better test-time scaling properties and effectively alleviates the ``overthinking" issue, establishing a solid baseline and offering new perspectives for future research.
\end{itemize}

\begin{figure*}[t!]
\vskip -0.1in
    \centering
    \includegraphics[width=0.95\linewidth]{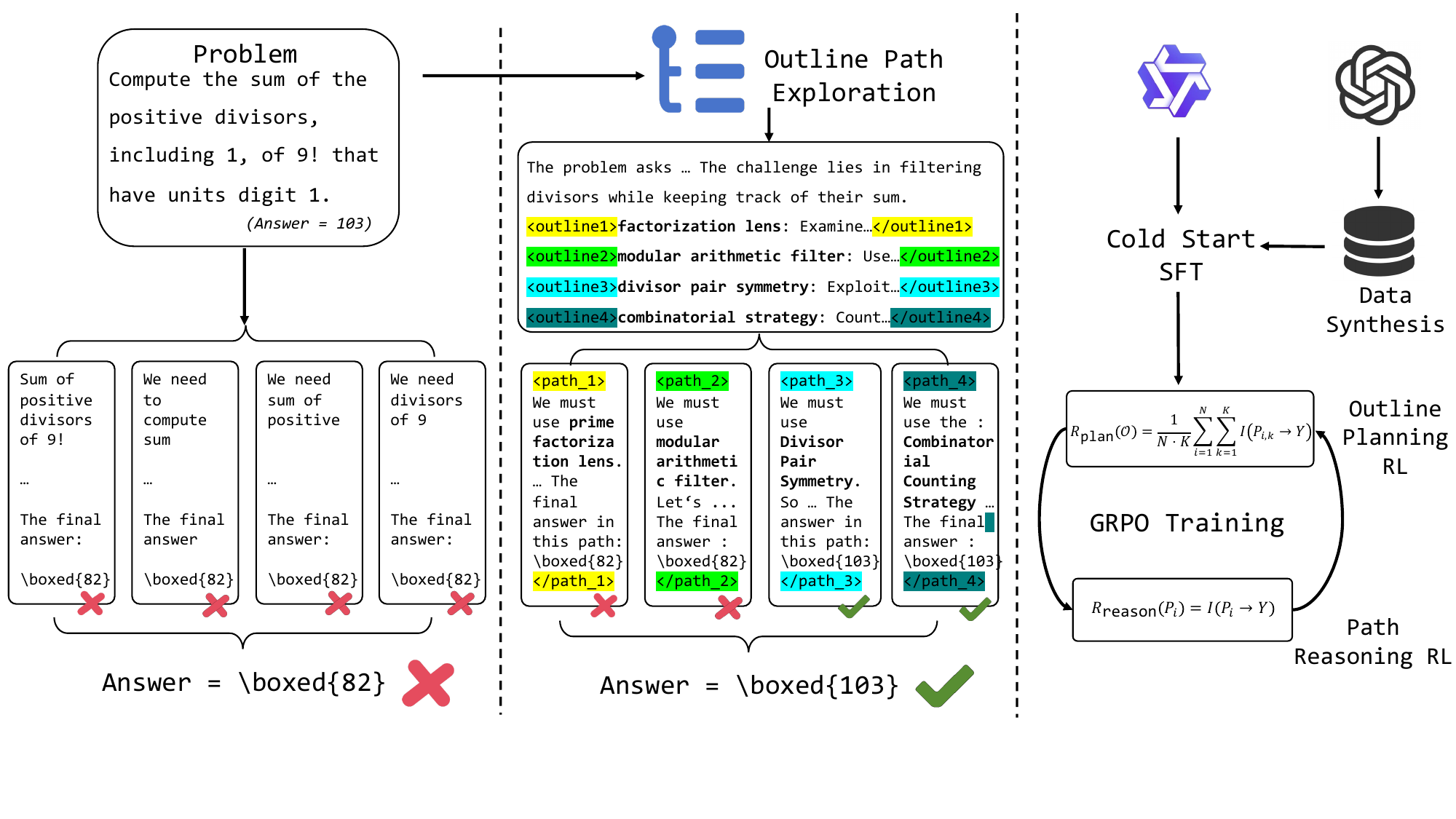}
    \vskip -0.45in
    \caption{\textbf{Overview of OPE framework.} 
\textbf{(Left)} Naive parallel thinking samples reasoning paths independently. Due to mode collapse, these paths often exhibit high redundancy and tend to converge on the same incorrect answer. 
\textbf{(Middle)} OPE mitigates this by explicitly generating diverse outlines to partition the solution space into distinct directions (four different strategies). This structured exploration maximizes the coverage of potential solutions, enabling the model to successfully locate the correct reasoning trajectory. 
\textbf{(Right)} The OPE training pipeline consists of a \textit{Cold Start} phase using synthesized data, followed by a novel \textit{Iterative RL} strategy.}
    \label{fig:framework}
    \vskip -0.1in
\end{figure*}

\section{Related Work}
\subsection{Parallel Thinking}
Parallel thinking typically involves generating multiple reasoning trajectories, followed by an aggregation phase to derive the final solution. The majority of existing research focuses on the aggregation phase, which can be broadly categorized into two streams. (1) Selection and Ranking Methods: These approaches evaluate individual paths via scoring or pairwise comparisons to identify the optimal solution. Early works employed Verifiers or Reward Models to assign scalar scores to reasoning steps or final answers, selecting the candidate with the highest confidence \citep{cobbe2021training, lightman2023let}. More recent methods utilize pairwise ranking mechanisms to filter out low-quality trajectories \citep{li2023making}. (2) Summary Methods: Instead of selecting a single path, these methods aim to synthesize a final answer by aggregating information from all generated paths. This includes heuristic strategies such as Self-Consistency, which relies on majority voting to determine the consensus answer \citep{wang2022self}, along with its various extensions. Other approaches leverage the inherent summarization capabilities of LRMs to merge diverse reasoning chains into a coherent conclusion \citep{jiang2023llm}. In contrast, the path exploration phase has received comparatively less attention. Skeleton of Thought \citep{sot} decomposes problems into sub-tasks to accelerate inference, and Leap \citep{leap} enhances efficiency through fixed information exchange mechanisms between paths. However, these methods are often constrained to complex architectural designs. Distinctly, OPE enables the model to explicitly determine the exploration distribution to identify potential solutions.
\vskip -0.1in
\subsection{RLVR}

RLVR leverages verifiable outcome signals, such as rule-based answer matching or model-based evaluation, to guide model optimization, thereby circumventing the reliance on human annotation for reward signals. This paradigm has demonstrated remarkable efficacy in enhancing the reasoning capabilities of Large Language Models (LLMs) \citep{trinh2024solving, shao2024deepseekmath}. 
Among a diverse array of RLVR algorithms, Group Relative Policy Optimization (GRPO) \citep{shao2024deepseekmath} eliminates the need for a separate value network and reduces training instability by utilizing this group-relative baseline and demonstrates significant advantages in mathematical reasoning tasks. Specifically, for a group of $G$ outputs $\{o_1, \dots, o_G\}$ sampled for a query $q$, GRPO estimates the advantage of each output $o_i$ as $A_i = (r_i - \text{mean}(\mathbf{r})) / \text{std}(\mathbf{r})$, where $\mathbf{r}$ is the vector of rewards. Building on this foundation, several variants have been proposed to further refine sample efficiency and stability \citep{vapo, gspo, dapo}. Despite the extensive exploration of RL in the reasoning scenarios, the application of RL specifically to parallel thinking remains underexplored. This paper bridges this gap by providing a theoretical analysis of parallel thinking and offer empirical insights for future research.

\section{Parallel Thinking with RLVR}
\label{sec:theory}

In this section, we present a theoretical analysis of the parallel thinking paradigm. We focus on the RLVR settings and identify that Mutual Information Saturation among generated paths during the exploration phase as a fundamental bottleneck that constrains the reasoning performance.
\subsection{Formulation of Parallel Thinking}
\label{formulation}
Let $Q$ denote the input query drawn from the data distribution $\mathcal{D}$, and $Y$ represent the corresponding ground-truth Answer. We define $A$ as the answer generated by a unified policy model $\pi_\theta$. The reasoning workflow in parallel thinking can be formally decomposed into two distinct phases:

1.  \textbf{Exploration:} Given $Q$, the model samples a set of $N$ parallel reasoning paths $\mathcal{P} = \{P_1, \dots, P_N\}$ from the policy:
    \vskip -0.2in
    \begin{equation}
    P_i \sim \pi_\theta(\cdot | Q), \quad \forall i \in \{1, \dots, N\}
    \end{equation}
    \vskip -0.1in

2.\textbf{Aggregation:} Based on the query $Q$ and the set of explored paths $\mathcal{P}$, the model synthesizes the final answer $A$:
    \vskip -0.2in
    \begin{equation}
    A \sim \pi_\theta(\cdot | \mathcal{P}, Q)
    \end{equation}
    \vskip -0.1in
We formulate the parallel paths 
\(\mathcal{P}\) as intermediate variables that bridge the query and the final answer. Consequently, the overall inference probability is derived by marginalizing over the space of generated paths:
\begin{equation}
\pi_\theta(A | Q) = \sum_{\mathcal{P}} \pi_\theta(A | \mathcal{P}, Q) \pi_\theta(\mathcal{P} | Q)
\end{equation}    

\subsection{Parallel Thinking Optimization}

The fundamental objective of RLVR is to optimize the policy parameters $\theta$ to maximize the expected reward over the data distribution $\mathcal{D}$:
\begin{equation}
J(\theta) = \mathbb{E}_{Q \sim \mathcal{D}} \left[ \mathbb{E}_{A \sim \pi_\theta(\cdot|Q)} \left[ r(Q, A) \right] \right]
\end{equation}
where $r(Q, A)$ serves as the reward function. In the RLVR setting, this reward is usually deterministic and binary (i.e., $r=1$ if the generated answer $A$ matches the ground truth $Y$, and $0$ otherwise). Consequently, maximizing the expected reward $J(\theta)$ is asymptotically equivalent to maximizing the log-likelihood of generating the ground truth $Y$ given $Q$:
\begin{equation}
\max_\theta J(\theta) \iff \max_\theta \log \pi_\theta(Y | Q)
\end{equation}
\vskip -0.1in
Given that the generation of the ground truth $Y$ depends on the intermediate reasoning paths $\mathcal{P}$, the log-likelihood can be expressed as $\log \sum_{\mathcal{P}} \pi_\theta(Y|\mathcal{P}, Q)\pi_\theta(\mathcal{P}|Q)$. Since directly optimizing the resulting log-sum is intractable, we follow standard practice and consider its Evidence Lower Bound (ELBO):
\begin{equation}
\log \pi_\theta(Y | Q) \ge \mathbb{E}_{\mathcal{P} \sim \pi_\theta(\cdot|Q)} \left[ \log \pi_\theta(Y | \mathcal{P}, Q) \right]
\end{equation}

From the perspective of Information Theory \citep{reza1994information_theory}, the expectation term $\mathbb{E}_{\mathcal{P}}[\log \pi_\theta(Y|\mathcal{P}, Q)]$ corresponds to the negative Conditional Entropy, denoted as $-H(Y | \mathcal{P}, Q)$. Recall the definition of Mutual Information (MI):
\begin{equation}
I(\mathcal{P}; Y | Q) = H(Y | Q) - H(Y | \mathcal{P}, Q)
\end{equation}
Since the ground truth $Y$ is deterministic for a given query $Q$ in the RLVR setting, the entropy $H(Y|Q)$ remains constant ($Y$ is not a random variable given $Q$). Consequently, maximizing the lower bound $\mathcal{L}_{\text{ELBO}}$ is mathematically equivalent to maximizing the mutual information:
\begin{equation}
\max_\theta \mathcal{L}_{\text{ELBO}} \iff \max_\theta I(\mathcal{P}; Y | Q)
\end{equation}
This derivation reveals that the optimization of parallel thinking is fundamentally to generate reasoning paths $\mathcal{P}$ that maximize the information gain regarding the ground-truth.

\subsection{Mutual Information Saturation among Paths}

Instead of analyzing the mutual information between the set $\mathcal{P}$ and $Y$ directly, we can simplify the formulation by decomposing the total mutual information item into the sum of marginal contributions from each individual path using the chain rule:
\vskip -0.15in
\begin{equation}
I(\mathcal{P}; Y | Q) = \sum_{i=1}^N I(P_i; Y | P_{1:i-1}, Q)
\end{equation}
\vskip -0.1in
In the naive parallel thinking paradigm described in \cref{formulation}, paths are sampled independently and identically (i.i.d.). However, since LRMs typically undergo extensive post-training, they often exhibit the risk of mode collapse, which leads to significant semantic redundancy among sampled trajectories. Consequently, the marginal information gain diminishes rapidly:
\begin{equation}
I(P_i; Y | P_{1:i-1}, Q) \to 0 \quad \text{as} \quad i \to N
\end{equation}
\vskip -0.1in
We term this phenomenon \textbf{Mutual Information Saturation}. Intuitively, this implies that when tackling complex problems, the model tends to fall into repetitive failure modes, where a large number of parallel paths converge to serveral incorrect answers rather than exploring diverse regions of the solution space. As a result, simply scaling the number of paths $N$ yields diminishing returns.

\begin{figure}
    \centering
    \includegraphics[width=1.0\linewidth]{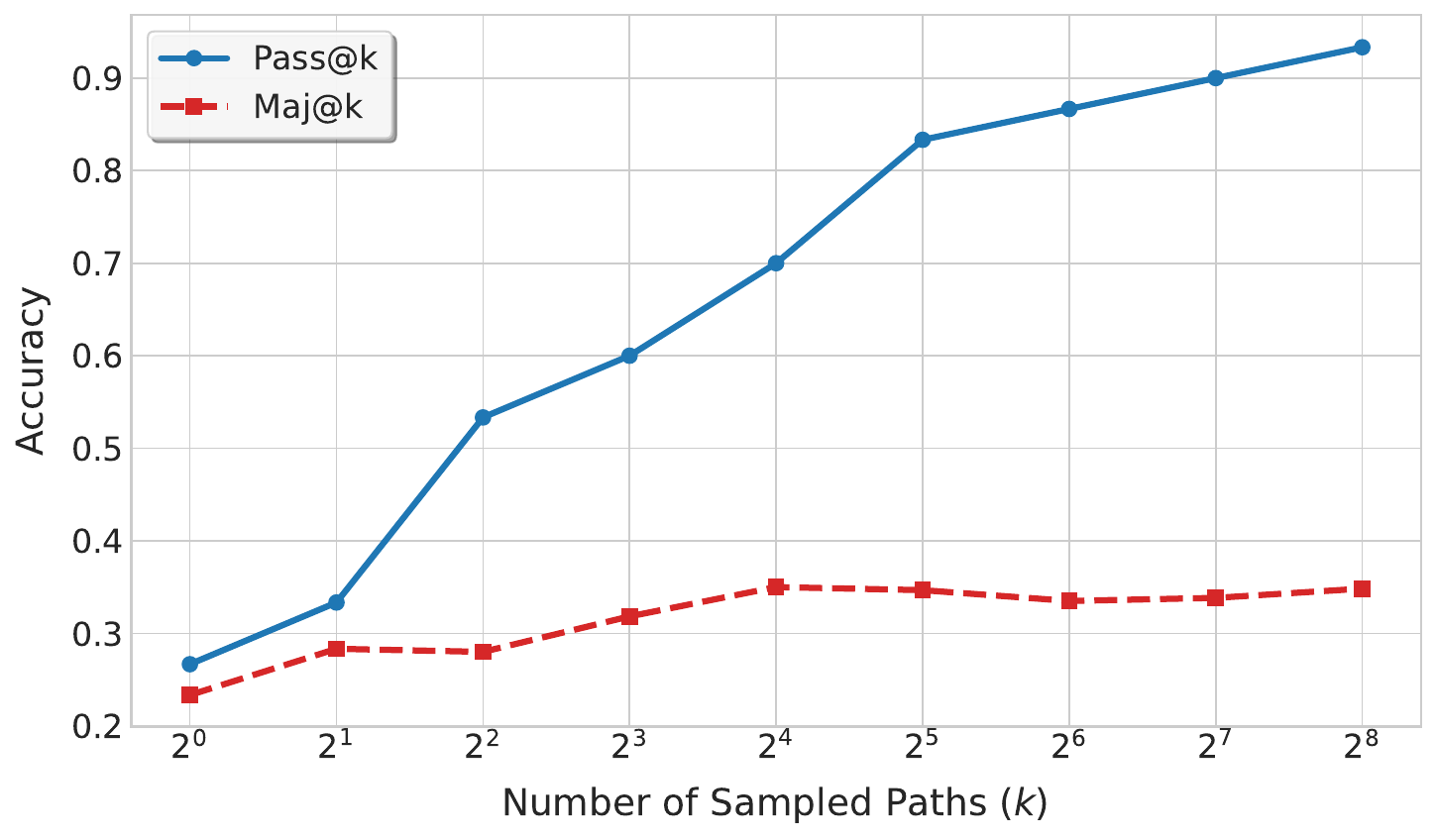}
    \caption{
Comparisons of Pass@k curves and Maj@k curves on the HMMT-25 benchmark.
}   
    \label{fig:preliminary}
\vskip -0.2in
\end{figure}

To empirically validate this phenomenon, we conduct experiments on the challenging mathematical benchmark HMMT-25 \citep{hmmt}. Specifically, we perform $N=256$ independent samplings for each query using DeepSeek-R1-Distill-Qwen-7B \citep{deepseek-r1} and analyze the scaling trends of two key metrics: Pass@k, which measures the probability that at least one path in $k$ samples is correct, and Maj@k, which represents the accuracy of the answer derived from majority voting over $k$ samples.
The results showed in \cref{fig:preliminary} reveal a critical discrepancy: while Pass@k continues to rise as the sample size increases, Maj@k plateaus after approximately 20 samples. This indicates that although the model possesses the intrinsic capability to solve the problems (evidenced by the rising Pass@k), the vast majority of sampled paths converge to incorrect answers. Consequently, the correct signal is drowned out by redundant errors, thereby imposing a hard ceiling on the performance of aggregation.

\section{Methodology}
\label{sec:method}
To mitigate the bottleneck of Mutual Information Saturation, we propose Outline-Guided Path Exploration (OPE), which introduces a set of explicit reasoning outlines during the exploration process to dynamically partition the solution space. In this section, we first analyze how the introduction of outlines theoretically overcomes the saturation bottleneck. Subsequently, we detail the practical implementation of OPE, which involves a cold-start stage followed by an iterative RL strategy.

\subsection{Breaking Saturation with OPE}
Under the naive parallel thinking paradigm, paths are sampled i.i.d., making it intrinsically difficult to optimize the marginal information gain $I(P_i; Y | P_{1:i-1}, Q)$. To address this, it is desirable to rationally partition the solution space to maximize the coverage of potential solutions before conducting concrete path exploration. OPE achieves this by guiding the model to explicitly plan the directions of exploration in the form of a set of distinct outlines $\mathcal{O} = \{O_1, \dots, O_N\}$ before executing reasoning steps. Formally, the exploration phase \(\pi_\theta(\mathcal{P}| Q)\) is reformulated as a hierarchical process:
\vskip -0.1in
\begin{equation}
\pi_\theta(\mathcal{P}, \mathcal{O} | Q) = \pi_\theta(\mathcal{O} | Q) \cdot \prod_{i=1}^N \pi_\theta(P_i | O_i, Q)
\end{equation}
\vskip -0.1in
$\mathcal{O}$ serve as structural directives, guiding the subsequent parallel generation towards $N$ distinct directions, each $O_i$ corresponding to a different strategy for solving the problem. \cref{fig:framework} illustrates the structural differences between OPE and naive parallel exploration.

Under this new paradigm, the original optimization objective   transforms into maximizing the joint mutual information $I(\mathcal{P}, \mathcal{O}; Y | Q)$. Using the chain rule, this objective can be orthogonally decomposed into two components:

\begin{equation}
I(\mathcal{P}, \mathcal{O}; Y | Q) = \underbrace{I(\mathcal{O}; Y | Q)}_{\text{Planning Gain}} + \underbrace{I(\mathcal{P}; Y | \mathcal{O}, Q)}_{\text{Reasoning Gain}}
\end{equation}

\label{objective}
\textbf{1. Planning Gain ($I(\mathcal{O}; Y | Q)$):} This term measures the extent to which the set of outlines $\mathcal{O}$ covers the solution space containing the ground truth. Since outlines are significantly shorter than full reasoning paths, they can be generated as a coherent sequence, enabling the model to explicitly manage diversity among outlines and avoid the diminishing returns associated with i.i.d. sampling.

\textbf{2. Reasoning Gain ($I(\mathcal{P}; Y | \mathcal{O}, Q)$):} This term quantifies the model's ability to generate correct reasoning trajectories given the constraints of the outlines. 

OPE aims to effectively transform the optimization of the total mutual information $I(\mathcal{P}, \mathcal{O}; Y | Q)$ into the optimization of the two objectives independently.
\subsection{OPE with RLVR}
\label{ope_training}
In this section, we describe the practical implementation of OPE on Qwen3-8B-Base \citep{qwen3} (the state-of-the-art base model in 8B size). The pipeline begins with a cold-start phase to instill the OPE reasoning pattern into the model. Subsequently, we present RL training tailored to the two optimization objectives analyzed in \cref{objective} using GRPO \citep{shao2024deepseekmath}, and propose an iterative training strategy to further enhance the performance.

\textbf{Data} Given that the primary goal of parallel thinking is to address complex reasoning tasks, we utilize the mathematical training corpus in GURU \citep{guru}, a high-quality dataset derived from OR1 \citep{or1}, DAPO \citep{dapo}, and DeepScaler \citep{deepscaler} sources after deduplication and difficulty filtering. In this paper, we specifically employ its subsets from OR1 and DAPO (36k samples in total), excluding DeepScaler data due to its relatively lower difficulty.

\textbf{Cold Start}  Inspired by Parallel-R1 \citep{parallel-r1}, which demonstrates that zero-shot trajectory synthesis is more stable on simpler queries, we select a subset of 5.4k queries from the full dataset where Qwen2.5-7B-Math \citep{team2024qwen2} achieves a pass rate greater than 0.5 for our cold-start data synthesis. We then prompt gpt-oss-20b \citep{oss} to first generate a brief analysis of the query, followed by $N$ distinct reasoning outlines in the tags \textless outline\_i$>\dots$\textless/outline\_i$>$. We then concatenate the generated outlines along with a path token $<$path\_i$>$ to the original query and require model to generate a corresponding reasoning path strictly following the guidance of $O_i$. Considering the context window constraints and the complexity of the reasoning tasks, we set $N=4$ and limit the maximum generation length per query to 8k tokens. To ensure a fair comparison with naive parallel thinking, we instruct gpt-oss-20b to directly generate reasoning paths for these queries, thereby constructing a cold-start dataset of identical size for naive parallel thinking.

\textbf{Outline Planning RL} This stage aims to maximize the Planning Gain $I(\mathcal{O}; Y | Q)$ described in \cref{objective}, which is equivalent to maximizing the expected reward of the generated outlines $\mathcal{O}$. However, unlike complete reasoning paths, outlines serve as high-level strategic plans and do not contain full problem resolutions, making them unsuitable for direct evaluation via verifiable reward functions. To address this, we estimate the quality of $\mathcal{O}$ by the success rate of the downstream reasoning paths generated following its guidance. Specifically, for each outline $O_i \in \mathcal{O}$, we append the $<$path\_i$>$ token and sample $K$ distinct reasoning paths, resulting in a total of $N \times K$ trajectories. The reward function for $\mathcal{O}$ is then defined as the average accuracy of these generated paths:
\vskip -0.1in
\begin{equation}
R_{\text{plan}}(\mathcal{O}) = \frac{1}{N \cdot K} \sum_{i=1}^{N} \sum_{k=1}^{K} \mathbb{I}(P_{i,k} \to Y)
\end{equation}
\begin{table*}[t]
    \centering
    \small
    \setlength{\tabcolsep}{8pt} 
    \renewcommand{\arraystretch}{1.15} 
    
    \caption{
Main results across various benchmarks under both Cold Start and RL phases. SC denotes Self-Consistency aggregation, while LRM refers to summary aggregation. The best results are highlighted in \textbf{bold}, and the second best results are \underline{underlined}.
}
    \label{tab:main_results}
    
    \begin{tabular}{llccccccc}
        \toprule
        \multirow{2}{*}{\textbf{Exploration}} & \multirow{2}{*}{\textbf{Aggregation}} & \multicolumn{7}{c}{\textbf{Datasets}} \\
        \cmidrule(lr){3-9}
         &  & \textbf{Math500} & \textbf{AMC}&   \textbf{AIME 24} & \textbf{AIME 25} & \textbf{HMMT-25}&\textbf{BeyondAIME} & \textbf{AVG.} \\
        \midrule
        
        \rowcolor{gray!10} 
        \multicolumn{9}{l}{\textit{SFT (Cold Start)}} \\ 
        
        \multirow{4}{*}{Naive} 
         & Random    & 78.35 & 58.94 & 16.43 & 17.11 & 6.56 & 4.58 & 30.33 \\
         & SC & 79.80 & 60.25 & 15.21 & 17.92 & 5.21 & 3.80 & 30.36 \\
         & BoN       & 90.00 & 80.75 & 30.63 & 28.54 & 13.12 & 10.60 & 42.27 \\
         & LRM       & \textbf{93.40} & \textbf{87.00} & \textbf{41.88} & \textbf{33.12} & \textbf{17.08} & \textbf{16.30} & \textbf{48.13} \\
         
        \cmidrule(lr){1-9} 
        
        \multirow{4}{*}{OPE} 
         & Random    & 74.55 & 53.37 & 15.81 & 13.59 & 3.67 & 4.58 & 27.60 \\
         & SC & 77.80 & 55.00 & 17.71 & 14.27 & 4.48 & 5.10 & 29.06 \\
         & BoN       & 88.00 & 72.75 & 29.27 & 23.44 & 9.69 & 10.60 & 38.96 \\
         & LRM       & \underline{90.40} & \underline{81.75} & \underline{38.23} & \underline{29.06} & \underline{14.06} & \underline{15.80} & \underline{44.88} \\
        
        \rowcolor{gray!10} 
        \midrule
        \multicolumn{9}{l}{\textit{RL Approaches}} \\ 
        
        \multirow{4}{*}{Naive} 
         & Random    & 83.75 & 66.81 & 23.54 & 22.03 & 9.71 & 7.40 & 35.54 \\
         & SC & 86.00 & 69.00 & 23.65 & 24.58 & 8.75 & 7.70 & 36.61 \\
         & BoN       & 91.00 & 85.25 & 43.12 & 32.29 & 16.15 & 15.20 & 47.17 \\
         & LRM       & 91.40 & 86.00 & 45.42 & \textbf{36.15} & 17.40 & 16.50 & 48.81 \\
         
        \cmidrule(lr){1-9}
        
        \multirow{4}{*}{OPE} 
         & Random    & 86.65 & 71.12 & 29.90 & 21.46 & 10.47 & 11.28 & 38.48 \\
         & SC & 90.20 & 75.75 & 31.67 & 22.81 & 11.25 & 11.40 & 40.51 \\
         & BoN       & \textbf{93.80} & \underline{87.75} & \underline{47.50} & \underline{34.69} & \underline{19.17} & \textbf{20.40} & \underline{50.55} \\
         & LRM       & \underline{92.20} & \textbf{88.75} & \textbf{49.17} & 34.17 & \textbf{20.62} & \underline{19.70} & \textbf{50.77} \\
        
        \bottomrule
    \end{tabular}
\end{table*}
where $P \to Y$ denotes that path $P$ successfully reaches the ground-truth answer $Y$. This formulation encourages the model to generate outlines that lead to high-probability success regions in the solution space.

\textbf{Path Reasoning RL} This phase aims to optimize $I(\mathcal{P}; Y | \mathcal{O}, Q)$. Since each path $P_i$ represents a complete reasoning trajectory, its correctness can be directly verified against the ground truth. Specifically, we optimize the generation of paths conditioned on a given outline $O_i$. Using the concatenation of the query, the outline, and the path token $<$path\_i$>$ as input, we employ the verifiable outcome as the reward function:
\begin{equation}
R_{\text{reason}}(P_{i}) = \mathbb{I}(P_{i} \to Y)
\end{equation}
Formally, the RL training in this phase differs from standard GRPO only by the inclusion of the $<$path\_i$>$ token. Therefore, to make fair comparisons, we train the naive parallel baseline using GRPO with the same reward function.

\textbf{Iterative OPE Training.} As indicated by the optimization objectives above, the training for outline planning and path reasoning are formally independent. Intuitively, the effectiveness of an outline is bounded by the problem-solving capability of models, while the reasoning performance is conversely dependent on the quality of the guiding outlines. Motivated by this interdependence, we adopt an iterative optimization strategy to achieve co-optimization of outline planning and path reasoning capabilities. In practice, we first conduct Outline Planning RL for a fixed number of steps. We then utilize the final checkpoint to generate outlines for training data, which serves as the input for the subsequent Path Reasoning RL phase. This cycle is repeated multiple times to achieve further improvements. Algorithm \ref{alg:ope} illustrates the complete workflow of the OPE training.

\section{Experiments}

\subsection{Setup}

\textbf{Baselines.} We select Naive Parallel Thinking as our primary baseline. This baseline is established with the same size cold-start data synthesized, followed by standard GRPO training, which is described in \cref{ope_training}. To analyze how the exploration of paths impacts the subsequent aggregation phase, we select four distinct aggregation strategies:
\begin{itemize}[itemsep=5pt, parsep=0pt, topsep=0pt, partopsep=0pt]
    \item \textbf{Random}: Randomly selects one generated path and uses its output as the final answer.
    \item \textbf{Self-Consistency (SC)} \citep{self-conf}: Determines the final answer via majority voting across all generated paths.
    \item \textbf{Best-of-N (BoN)}: Selects the correct answer if at least one path in the set is correct. This serves as the theoretical upper bound of the exploration phase.
    \item \textbf{LRMs-Based Summary}: Prompts the LRM to reflect on the generated paths and synthesize a final answer.
\end{itemize}

\textbf{Training Recipes.} For the cold-start phase, we conduct supervised fine-tuning with lr = 1e-5 for 2 epochs. The subsequent RL training is conducted using the VeRL framework \citep{verl}, with a batch size of 256 and a learning rate fixed at 1e-6. We randomly sample 2k queries from the 36k dataset as a validation set, utilizing the remaining queries for RL training. For all RL stages, we set the rollout size to 8. In outline planning RL, each aspect $O_i$ of the outlines $\mathcal{O}$ is randomly paired with $K = 4$ sampled paths to obtain stable reward estimates. Following the practice in \citep{guru}, the naive RL training is conducted for 2 epochs (270 steps in total). To align the OPE training at the step level, we train the Outline Planning RL for 70 steps, then switch to Path Reasoning RL for 65 steps, and repeat this cycle twice to match the total training steps. Detailed training parameters are listed in Appendix \ref{app:training}.

\textbf{Evaluation.} We selected six benchmarks with various difficulty including MATH-500 \citep{math500}, AMC23, AIME24 \& 25, BeyondAIME \citep{bytedance_seed_2025_beyondaime}, and HMMT-25 \citep{hmmt}. We duplicated the queries in AMC23 and BeyondAIME 10 times, and those in AIME24 \& 25, and HMMT-25 32 times to ensure statistically stable evaluations. We report accuracy as the primary metric, employing the math-verify library \citep{math_verify} to rigorously assess the equivalence between the predictions and the ground truth. Regarding the aggregation strategies, when a tie occurs in Self-Consistency, we select the answer with the longer average output length. For LRMs-Based Summary, we employ the Qwen3-8B \citep{qwen3} with a generic summary prompt. More details of our evaluation are provided in Appendix \ref{app:eval}.

\subsection{Main Results}

Table \ref{tab:main_results} presents the comparison between Naive and OPE exploration strategies across various aggregation methods and benchmarks. The results yield several key insights:

\textbf{OPE consistently outperforms naive baselines.}
While OPE initially underperforms Naive exploration in the Cold Start phase, which may be attributed to the domain shift introduced by the outline-guided reasoning format, it demonstrates superior scalability and adaptability during RL training. After RL, OPE achieves the highest average performance across all aggregation methods. Notably, OPE yields the most significant gains in Self-Consistency (SC) aggregation, improving the average accuracy from 36.61\% to 40.51\%. This improvement indicates that OPE effectively alleviates the bottleneck in majority voting (\cref{fig:preliminary}) caused by mutual information saturation.

\textbf{OPE demonstrates superior improvements on challenging tasks.}
The performance advantage of OPE becomes increasingly pronounced as task difficulty increases. While the improvement on the relatively simpler MATH-500 dataset is moderate (93.80\% vs. 91.00\% using BoN), the gains on the more challenging BeyondAIME benchmark are substantial, with OPE achieving 20.40\% accuracy in BoN compared to the Naive baseline's 15.20\% (+5.20\%). This trend suggests that by structuring the exploration process, OPE helps the model find correct paths that have low probabilities and are often missed by simple independent sampling.

\textbf{OPE breaks the information bottleneck for LRMs-Based summary.} Different the other three rule-based aggregation method that rely solely on the extracted answer in each path, LRMs-based summary can analyze the detailed information within each generated path and perform holistic reasoning over all candidates. 
While naive parallel thinking benefits from RL training under rule-based aggregation strategies, its improvement under the LRMs-Based Summary approach remains limited (from 48.13\% to 48.81\%). This suggests that in the naive setting, mutual information saturation among paths restricts the amount of new knowledge available for LRM aggregation. 
In contrast, with OPE, the average accuracy increases from 44.88\% to 50.77\%. These results demonstrate that the advantage of OPE lies not only in increasing the probability of discovering correct answers, but also in providing the summary model with a more diverse set of reasoning knowledge and informative traces within the explored paths, enabling more effective aggregation. 

\textbf{OPE expands the frontier of path explorations.}
The BoN metric serves as a proxy for the upper bound of the exploration phase. OPE consistently achieves the highest BoN scores across all benchmarks after RL training, improving the average accuracy by 3.38\% (50.55\% vs. 47.17\%). Note that the parallel sampling budget is constrained to $N=4$, which highlights the models' outline planning capability of identify high-probability directions within the solution space, showcasing its potential in low-resource scenarios.
\begin{table}[t]
    \centering
    \small
    \setlength{\tabcolsep}{6pt}
    \renewcommand{\arraystretch}{1.1}
    \caption{Average results across benchmarks of different stages.}
    \label{tab:ablation_iterative}
    \begin{tabular}{lcccc}
        \toprule
        \textbf{Stage} & \textbf{Rand} & \textbf{Self} & \textbf{BoN} & \textbf{LRM}\\
        \midrule
        Cold Start & 27.60 & 29.06 & 38.96 & 44.88\\
        \midrule
        \textit{Iter 1} & & & \\
        + Outline RL & 31.82 & 32.55 & 43.61 & 48.41\\
        + Path RL    & 37.03 & 38.22 & 49.18 & 50.11\\
        \midrule
        \textit{Iter 2} & & & \\
        + Outline RL & 37.32 & 38.90 & 49.97 & 50.47\\
        + Path RL & \textbf{38.48} & \textbf{40.51} & \textbf{50.55} & \textbf{50.77}\\
        \bottomrule
    \end{tabular}
     \vskip -0.15in
\end{table}

\begin{figure}[b]
 \vskip -0.15in
    \centering
    \includegraphics[width=1.0\linewidth]{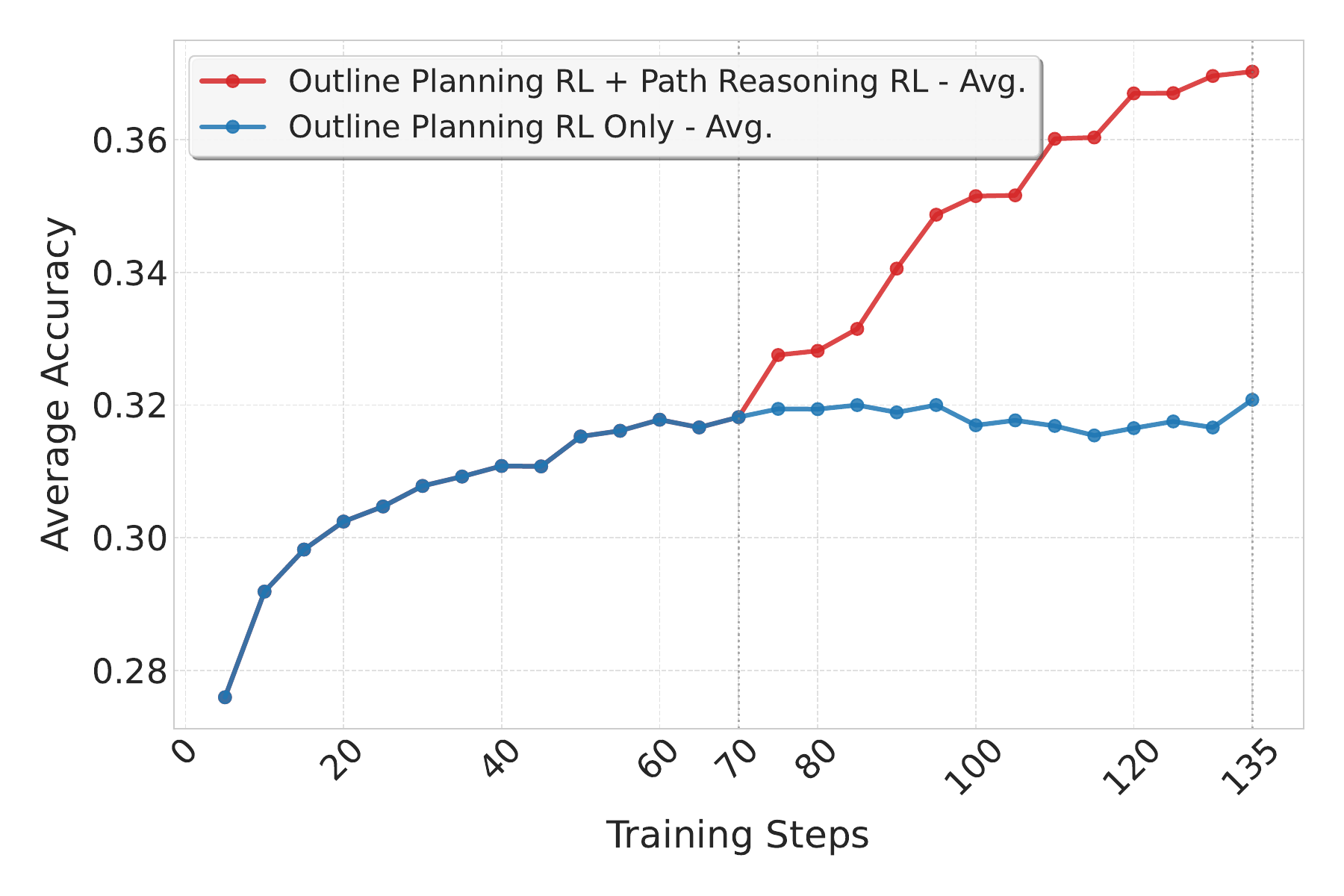}
    \vskip -0.1in
    \caption{
Comparison between continued Outline Planning RL (from 70 steps) and Path Reasoning RL, evaluated using average accuracy across all benchmarks with Random aggregation.
}
    \label{fig:ablation}
    \vskip -0.1in
\end{figure}
\subsection{Ablations}

To validate the effectiveness of our proposed iterative co-optimization strategy, we conduct ablation studies focusing on the incremental gains from each training stage. 
Table \ref{tab:ablation_iterative} reports the average accuracy across all benchmarks for each phase. The initial Outline RL significantly expands the exploration frontier, while the subsequent Path RL effectively converts this potential into reliable execution. The second iteration yields further gains across all metrics, confirming that outline planning and path reasoning are mutually reinforcing: better outlines guide more effective reasoning, which in turn enables the validation of more complex outlines.
We further examine whether the performance gains achieved in the Path Reasoning RL phase result from genuine improvements in path optimization, as opposed to merely benefiting from additional training steps. To this end, we extend the Outline Planning RL stage for an equal number of steps and compare its outcomes with those of Path Reasoning RL. As illustrated in \cref{fig:ablation}, Path Reasoning RL consistently yields superior results compared to the extended Outline Planning RL, confirming that the iterative strategy is crucial for fully realizing the potential of OPE.

\section{Analysis}
\subsection{Statistical Analysis}
\label{sec:stats}

While the main results demonstrate the overall effectiveness of OPE, we further analyze its impact from two perspectives: outline planning and path reasoning. From the outline perspective, we examine whether explicit outlines facilitate effective partitioning of the solution space. As shown in Table \ref{tab:stats}, OPE generates a greater number of unique answers (27.6 vs. 23.5), indicating broader and more diverse exploration compared to the naive baseline. This diversity suggests that OPE successfully mitigates mode collapse and guides the model to cover more distinct solution regions.

From the path reasoning perspective, we investigate whether outline-guided reasoning leads to clearer and more efficient problem solving. Results in Table \ref{tab:stats} demonstrate that OPE yields significantly more concise reasoning paths, with an average correct path length of 1,891 tokens—about 10\% shorter than the naive approach (2,217 tokens). This reduction in length implies that the model, when conditioned on well-structured outlines, can focus its reasoning and avoid redundant thinking patterns.

\begin{table}[t]
    \centering
    \small
    \setlength{\tabcolsep}{8pt}
    \caption{
Statistical comparison ($N=256$). Unique Ans denotes the average number of distinct answers per query, and Avg Length reports the average token count of correct reasoning paths.
}
    \label{tab:stats}
    \begin{tabular}{lcc}
        \toprule
        \textbf{Method} & \textbf{Unique Ans} & \textbf{Avg Tokens}  \\
        \midrule
        Naive & 23.5 &  2217.3 \\
        OPE   & \textbf{27.6} &  \textbf{1890.9 } \\
        \bottomrule
    \end{tabular}
    \vskip -0.1in
\end{table}

\subsection{Test-Time Scaling Analysis}
\label{sec:tts}

\begin{figure}[t]
    \centering
    \includegraphics[width=0.95\linewidth]{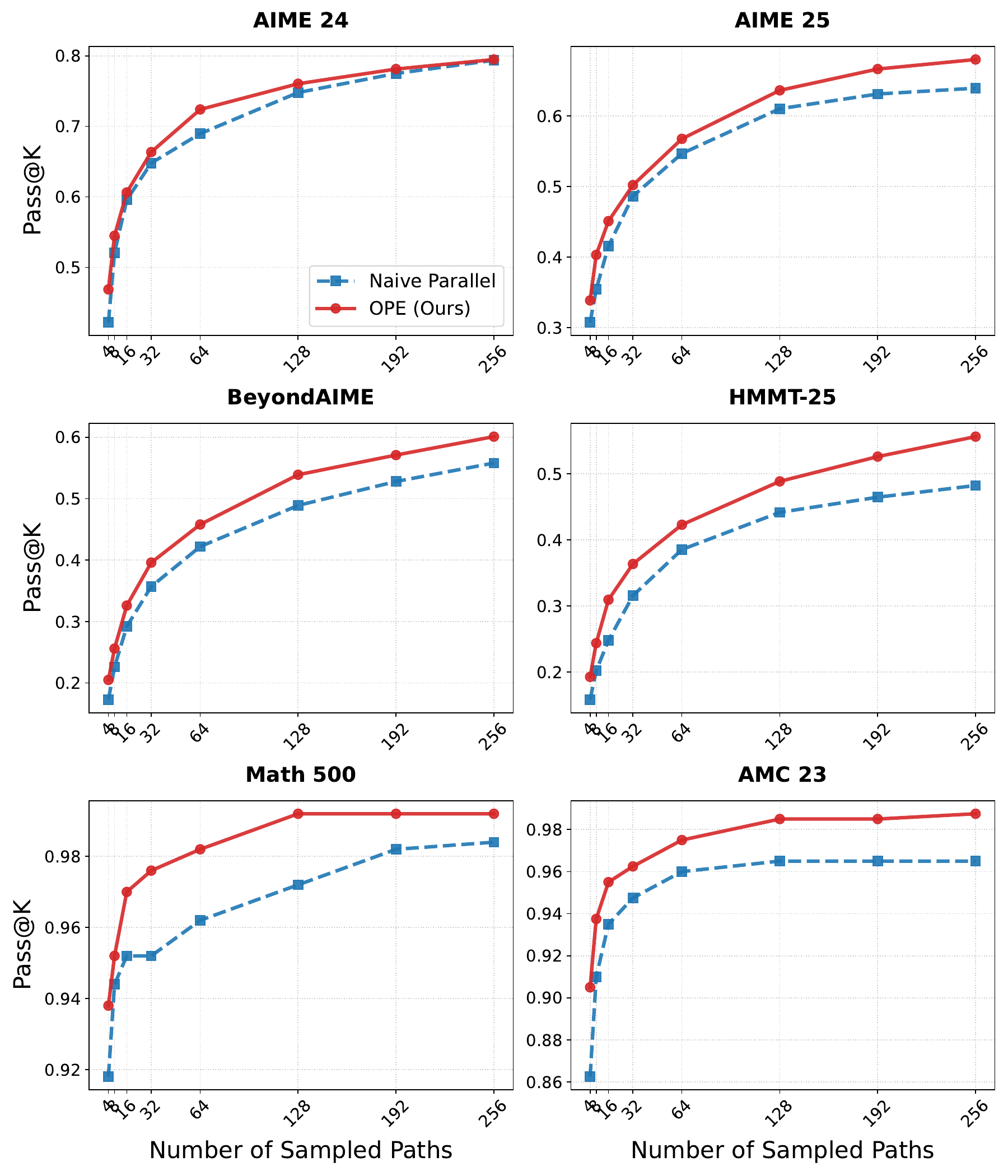}
    \caption{Pass@k scaling comparisons.}
    \label{fig:tts}
    \vskip -0.2in
\end{figure}

Although main results show that OPE effectively expands the reasoning frontier, as evidenced by the substantial improvements in BoN accuracy, it is constrained by a fixed outline budget (e.g., $N=4$ outlines per inference). 
To investigate whether OPE maintains its advantage under increased computational resources, we evaluate its scaling properties by performing multiple independent OPE samplings for each query. Figure \ref{fig:tts} illustrates the Pass@k scaling curves across different benchmarks. 
We observe that OPE exhibits superior scaling laws compared to the Naive baseline. 
For instance, on MATH-500 and AMC, OPE rapidly approaches nearly 100\% Pass@k as the sampling budget increases, while the naive approach saturates at a significantly lower performance ceiling. On more challenging benchmarks, the performance gap between OPE and naive continues to widen as the sample count increases, further highlighting OPE’s advantage in discovering difficult solutions through diverse exploration.
This trend validates its potential to scale with additional computational resources, allowing performance to continue improving without being limited by the initial outline budget.

\subsection{Failure Modes Analysis}
\label{sec:failure_analysis}
We conduct a analysis of query-level outcomes to understand the failure modes of naive and OPE strategies. We find that OPE successfully solves a substantially greater number of queries where the naive baseline fails (see Table~\ref{tab:cross-matrix}), demonstrating OPE's enhanced capability to discover correct solutions in regions of the solution space. To further analyze these failure modes, we count the frequency of correct reasoning paths for queries where only one method succeeds (see Table~\ref{tab:score1-case-study}). Notably, in “naive correct, OPE wrong” cases, the correct path typically appears only once (72\%), indicating these queries are solved mainly as a result of sampling stochasticity rather than robust reasoning. In contrast, for “OPE correct, naive wrong” instances, OPE is able to consistently produce multiple correct paths (about 40\% queries with more than 1 correct paths), highlighting the effectiveness of outline planning in reliably covering broader and more challenging portions of the solution space.

\section{Conclusion}
In this paper, we introduce Outline-guided Parallel Exploration (OPE), enhancing parallel thinking in LRMs by explicitly partitioning the solution space with diverse reasoning outlines and leveraging iterative reinforcement learning. Experiments show that OPE significantly enhances performance and scalability, especially for challenging tasks, by mitigating mutual information saturation and improving exploration, indicating its potential for advancing parallel reasoning in complex scenarios.

\newpage
\section*{Impact Statement}
This paper presents work whose goal is to advance the field of Large Reasoning Models. There are many potential societal consequences of our work, none
which we feel must be specifically highlighted here.

\nocite{langley00}

\bibliography{example_paper}
\bibliographystyle{icml2026}

\newpage
\appendix
\onecolumn

\begin{algorithm}
\small
  \caption{OPE Training}
  \label{alg:ope}
  \begin{algorithmic}[1]
    \STATE {\bfseries Input:} Dataset $\mathcal{D}$, Initial Model $\pi_{\theta}$, Iterations $M=2$, Steps $S_{\text{plan}}=70$, Steps $S_{\text{reason}}=65$
    \STATE {\bfseries Hyperparameters:} Outline Count $N$, Path Samples for Reward Estimation $K$
    \STATE {\bfseries Output:} Optimized Model $\pi_{\theta}$
    
    \FOR{$m=1$ {\bfseries to} $M$}
        \STATE \COMMENT{\textbf{Phase 1: Outline Planning RL (Optimize $\pi(\mathcal{O}|Q)$)}}
        \FOR{$s=1$ {\bfseries to} $S_{\text{plan}}$}
            \STATE Sample batch of queries $Q \sim \mathcal{D}$
            \STATE Generate outline sets $\mathcal{O} = \{O_1, \dots, O_N\} \sim \pi_\theta(\cdot | Q)$
            \FOR{each outline $O_i \in \mathcal{O}$}
                \STATE Append token \texttt{<path\_i>} and sample $K$ paths: $\{P_{i,k}\}_{k=1}^K \sim \pi_\theta(\cdot | Q, O_i)$
            \ENDFOR
            \STATE Compute Reward: $R_{\text{plan}}(O_i) = \frac{1}{K} \sum_{k=1}^{K} \mathbb{I}(P_{i,k} \to Y)$
            \STATE Update $\theta$ via GRPO to maximize $R_{\text{plan}}$
        \ENDFOR
        
        \STATE \COMMENT{\textbf{Transition: Construct Dataset for Phase 2}}
        \STATE Initialize $\mathcal{D}_{\text{reason}} \leftarrow \emptyset$
        \FOR{each query $Q \in \mathcal{D}$}
            \STATE Generate outlines $\mathcal{O} \sim \pi_\theta(\cdot | Q)$ using current policy
            \STATE $\mathcal{D}_{\text{reason}} \leftarrow \mathcal{D}_{\text{reason}} \cup \{(Q, O_i) \mid O_i \in \mathcal{O}\}$
        \ENDFOR
        
        \STATE \COMMENT{\textbf{Phase 2: Path Reasoning RL (Optimize $\pi(P|\mathcal{O}, Q)$)}}
        \FOR{$s=1$ {\bfseries to} $S_{\text{reason}}$}
            \STATE Sample batch $(Q, O_i) \sim \mathcal{D}_{\text{reason}}$
            \STATE Construct Inputs: $X = [Q, O_i, \texttt{<path\_i>}]$
            \STATE Sample paths: $P \sim \pi_\theta(\cdot | X)$
            \STATE Compute Reward: $R_{\text{reason}}(P) = \mathbb{I}(P \to Y)$
            \STATE Update $\theta$ via GRPO to maximize $R_{\text{reason}}$
        \ENDFOR
    \ENDFOR
    \STATE \textbf{return} $\pi_\theta$
  \end{algorithmic}
\end{algorithm}
\begin{table}[ht]
    \centering
    \small
    \setlength{\tabcolsep}{10pt}
    \caption{Cross-matrix statistics of query success: Naive (rows) vs. OPE (columns).}
    \label{tab:cross-matrix}
    \begin{tabular}{lcc}
        \toprule
        & OPE Correct & OPE Incorrect \\
        \midrule
        Naive Correct   & 1595 & 232 \\
        Naive Incorrect & 402  & 2551 \\
        \bottomrule
    \end{tabular}
\end{table}

\begin{table}[ht]
    \centering
    \small
    \setlength{\tabcolsep}{8pt}
    \caption{Distribution of correct path occurrences per query, split into “Naive correct, OPE wrong” and “OPE correct, Naive wrong” cases.}
    \label{tab:score1-case-study}
    \begin{tabular}{c|cc}
        \toprule
        \textbf{Frequency of correct paths (Total N=4 paths)} & \textbf{Naive correct, OPE wrong} & \textbf{OPE correct, Naive wrong} \\
        \midrule
        1 & 167 / 232 (71.98\%) & 243 / 402 (60.45\%) \\
        2 & 38 / 232 (16.38\%)  & 93  / 402 (23.13\%) \\
        3 & 17 / 232 (7.33\%)   & 40  / 402 (9.95\%)  \\
        4 & 10 / 232 (4.31\%)   & 26  / 402 (6.47\%)  \\
        \bottomrule
    \end{tabular}
\end{table}

\begin{table}[ht]
\centering
\caption{Training Hyperparameters}
\label{tab:training_params}
\begin{subtable}{0.4\textwidth}
\centering
\caption{Cold Start}
\begin{tabular}{lr}
\toprule
\textbf{Hyperparameter} & \textbf{Value} \\
\midrule
Epochs & 2 \\
Batch Size & 256 \\
Max Length & 8,192 \\
Learning Rate (LR) & $1 \times 10^{-5}$ \\
Minimum Learning Rate & $1 \times 10^{-7}$ \\
LR Decay Style & Cosine \\
LR Warmup Fraction & 0.1 \\
Adam Beta1 & 0.9 \\
Adam Beta2 & 0.95 \\
Weight Decay & $1 \times 10^{-2}$ \\
\bottomrule
\end{tabular}
\end{subtable}
\hfill
\begin{subtable}{0.55\textwidth}
\centering
\caption{RL configs using VERL.}
\begin{tabular}{lr}
\toprule
\textbf{Hyperparameter} & \textbf{Value} \\
\midrule
Epochs & 1 \\
data.train\_batch\_size & 256 \\
data.max\_prompt\_length & 2,048 \\
data.max\_response\_length & 8,192 \\
actor\_rollout\_ref.actor.optim.lr & $1 \times 10^{-6}$ \\
actor\_rollout\_ref.actor.ppo\_mini\_batch\_size & 32 \\
actor\_rollout\_ref.actor.ppo\_micro\_batch\_size\_per\_gpu & 4 \\
actor\_rollout\_ref.rollout.n & 8 \\
actor\_rollout\_ref.rollout.name & vllm \\
\bottomrule
\end{tabular}
\end{subtable}
\end{table}

\newpage

\section{Implementation Details}

\subsection{Data Construction}
\label{app:data}
To construct data for the cold-start phase, we leveraged the metadata provided in the GURU \citep{guru} dataset, which includes pass rate statistics evaluated by Qwen2.5-7B-Math. We specifically selected the subset of queries with a pass rate greater than 0.5, ensuring that the problems are solvable. Using these queries, we prompted gpt-oss-20b to generate data following the OPE format. The prompt used for synthesis is illustrated in Appendix \ref{cs_prompt}. To ensure the quality of the synthesized trajectories, we employed rejection sampling: for each query, we generated multiple candidate trajectories and retained only those that yielded the correct final answer.

\subsection{Training}
\label{app:training}
All training experiments, including both the Cold Start SFT and the Iterative RL phases, were conducted on a cluster of 32 NVIDIA H800 GPUs (configured as 4 nodes $\times$ 8 GPUs). The detailed hyperparameters for both stages are listed in Table \ref{tab:training_params}.

\subsection{Evaluation}
\label{app:eval}
During the evaluation phase, we adhered to the recommended settings for the Qwen3 model family, setting the sampling temperature to $0.6$ and top-p to 0.95. To ensure consistency with the training phase and accommodate complex reasoning chains, we set the maximum output token limit for each path to 8K tokens. We extract the final answer from the model's output by parsing the content within the last \texttt{\textbackslash boxed\{\}} command. The extracted answer is then compared against the ground truth using the \texttt{math-verify} \citep{math_verify} library to determine correctness.

\section{Prompts}

\subsection{System Prompt For Cold Start Data Synthesis \& OPE Training}
\label{cs_prompt}
\begin{tcolorbox}[
    listing only,
    listing style=promptstyle,
    colback=blue!5!white,
    colframe=blue!50!black,
    title=System Prompt For Cold Start Data Synthesis \& OPE Training,
    fonttitle=\bfseries,
    width=\textwidth,
    breakable,
    enhanced,
    drop shadow,
]
You are an advanced reasoning assistant capable of Outline-Guided Parallel Thinking.

Instructions:

1. Analysis \& Planning: Analyze the query and generate several distinct and diverse outlines representing different solving strategies. Each outline explores a different logical direction.

2. Independent Parallel Exploration: Generate reasoning paths strictly following each corresponding outline. Path $P_i$ must execute the strategy prescribed in Outline $O_i$ without deviation.
\end{tcolorbox}

\subsection{System Prompt LRMs-Based Summary}
\begin{tcolorbox}[colback=blue!5!white, colframe=blue!50!black, title=\textbf{System Prompt For LRMs-Based Summary}]
You are a great reasoner. 
Here is a problem, and multiple thinkers attempt to give their thought processes independently. 
Each thinker has written its own thought process towards the final answer. Each thinker is encouraged to take the other thinkers’ progress into account to reach the final answer.

\vspace{0.3cm}
\textbf{\# ====== Problem ======}

\{problem\}

\vspace{0.3cm}
\textbf{\# ====== Thinkers Thought Process ======}

\textbf{Thinker 1:}
\{path\_1\}

\textbf{Thinker 2:}
\{path\_2\}

\textbf{Thinker 3:}
\{path\_3\}

\textbf{Thinker 4:}
\{path\_4\}

\vspace{0.3cm}
Look at the above problem and thought process from each thinker, summarize from these thought processes and finally give your answer within \texttt{\textbackslash boxed}.
Summarize their thinking on the problem and try to summarize the thinking of these thinkers. Analyze the differences in thinking between these thinkers and try to analyze which thought process is correct. 

\vspace{0.2cm}
\textbf{Note:} 
\begin{itemize}
    \item It is generally believed that when most thinkers get the same answer, the answer may be correct. But you can't do it so superficially, because the correct answer may come from very few thinkers, or even no thinker gives the correct answer. For this reason, when you summarizing, you NEED adhere to the principles of professionalism and critical thinking, carefully identify these thought processes, and give a summary and final answer. 
    \item If you realize that none of these thinkers have answered correctly, you can even learn from the wrong experiences in the thought process of these thinkers and re-think the given problem to give the answer you think is most correct.
    \item Please DO NOT just solve the given problem independently like other thinkers, but summarize the thought process of all thinkers. In other words, you need to give the summary first, and then give the final answer within \texttt{\textbackslash boxed}, you can re-think this problem only if you realize that none of these thinkers have answered correctly.
\end{itemize}
\end{tcolorbox}

\section{Case Study}

\begin{tcolorbox}[colback=blue!5!white, colframe=blue!50!black, title=\textbf{Case Study}, breakable]

The problem asks for the sum of all positive divisors of $9!$ that end with the digit $1$. To solve it, we need to understand the structure of $9!$, identify which of its divisors satisfy the units‑digit condition, and then sum those divisors. Key considerations include the prime factorization of $9!$, the properties of divisors, and the behavior of units digits under multiplication. The solution space is finite but large, so an efficient strategy involves leveraging modular arithmetic and combinatorial counting of divisor pairs. The challenge lies in filtering divisors by their last digit while keeping track of their sum.

\texttt{<outline\_1>}**Prime Factorization Lens**
Examine the prime factorization of $9!$ to understand the building blocks of its divisors and how their units digits are determined.\texttt{</outline\_1>}

\texttt{<outline\_2>}**Modular Arithmetic Filter**
Use congruence modulo $10$ to identify which divisor combinations yield a units digit of $1$, narrowing the set of candidates.\texttt{</outline\_2>}

\texttt{<outline\_3>}**Divisor Pair Symmetry**
Exploit the fact that divisors come in pairs whose product is $9!$ to simplify the counting and summing process.\texttt{</outline\_3>}

\texttt{<outline\_4>}**Combinatorial Counting Strategy**
Count the number of divisors with the desired units digit by analyzing the exponents in the prime factorization and applying combinatorial principles.\texttt{</outline\_4>}

\texttt{<path\_1>}

We need to compute sum of positive divisors of 9! that have units digit 1. We must use prime factorization lens. So we need to factor 9!, find divisors, filter by units digit 1, sum. Let's do step by 
\ldots
$$
=4 \Rightarrow 2^03^4 = 81
$$

Both $1$ and $81$ are divisors of $9!$ (since $9!$ contains $3^4$).

**Step 7 – Compute the sum**
$$
1 + 81 = 82
$$

$$
\boxed{82}
$$
\texttt{</path\_1>}

\texttt{<path\_2>}

We need to compute sum of positive divisors of 9! that have units digit 1. We must use modular arithmetic filter. Let's proceed.

First compute 9! = 9*8*7*6*5*4*3*2*1. Calculate: 9! = 362880. But we c
\ldots
sidered all a,b,e combinations. The only ones that gave units digit 1 were the ones we found. Thus sum is 106.

Thus final answer: $\boxed{106}$. The final answer in this path: 
$$
\boxed{106}
$$
\texttt{</path\_2>}

\texttt{<path\_3>}

We need to compute sum of positive divisors of 9! that have units digit 1. We must use the outline: Divisor Pair Symmetry. So we should use the fact that divisors come in pairs whose product is 9!. We
\ldots
uct 42 $\rightarrow$ 2. Not 1. a=8,b=1,d=1: u2=6, u3=3, u7=7 $\Rightarrow$ product 126 $\rightarrow$ 6. Not 1.

Thus only 1 and 21.

Thus sum = 22.

We should output $\boxed{22}$. The final answer in this path: 
$$
\boxed{22}
$$
\texttt{</path\_3>}

\texttt{<path\_4>}

We need to compute sum of positive divisors of 9! that have units digit 1. We must use the outline: Combinatorial Counting Strategy. So we need to analyze prime factorization of 9!, determine which di
\ldots
$b=1,d=1$  $\rightarrow$ $d=21$

$b=4,d=0$  $\rightarrow$ $d=81$

Thus the divisors of $9!$ with units digit $1$ are $1,21,81$.

***3. Sum of these divisors***

$$
1+21+81=103.
$$

$$
\boxed{103}
$$
\texttt{</path\_4>}

\end{tcolorbox}


\begin{figure}[htbp]
    \centering
    \includegraphics[width=1.0\linewidth]{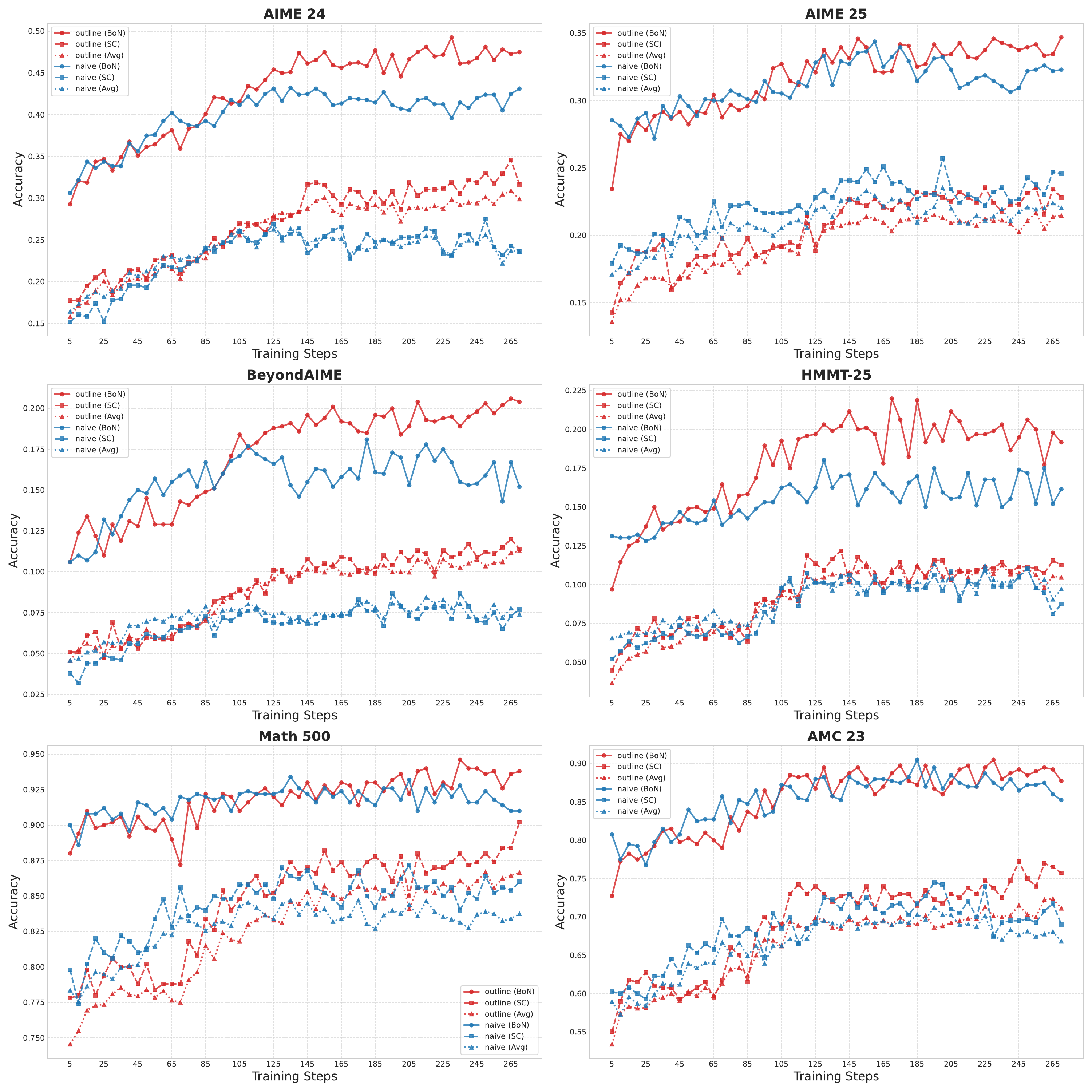}
    \caption{Comparison of the performance curves of Naive and OPE approaches across different datasets as training steps progress.}
    \label{fig:placeholder}
\end{figure}

\end{document}